\documentclass[10pt,twocolumn,letterpaper]{article}

\usepackage{subcaption}
\usepackage{graphicx} 
\usepackage{booktabs}
\usepackage[pagebackref,breaklinks,colorlinks]{hyperref}
\begin{document}
\title{Fashion Object Detection for Tops \& Bottoms}

\author{Andreas Petridis\thanks{Research undertaken whilst on internship at Levi Strauss \& Co. between September 2021 and February 2022}\\
Maastricht University, the Netherlands\\
{\tt\small a.petridis@student.maastrichtuniversity.nl}
\and
Mirela Popa\\
Maastricht University, the Netherlands\\
{\tt\small mirela.popa@maastrichtuniversity.nl}
\and
Filipa Peleja\\
Levi Strauss \& Co., Belgium\\
{\tt\small filipapeleja@gmail.com}
\and
Dario Dotti\\
Levi Strauss \& Co., Belgium\\
{\tt\small ddotti@levi.com}
\and
Alberto de Santos\\
Levi Strauss \& Co., Belgium\\
{\tt\small adesantos@levi.com}
}

% \date{May 2023}

\maketitle

\begin{abstract}
   Fashion is one of the largest world's industries and computer vision techniques have been becoming more popular in recent years, in particular, for tasks such as object detection and apparel segmentation. Even with the rapid growth in computer vision solutions, specifically for the fashion industry, many problems are far for being resolved \cite{cheng2021fashion}. Therefore, not at all times, adjusting out-of-the-box pre-trained computer vision models will provide the desired solution. In the present paper is proposed a pipeline that takes a noisy image with a person and specifically detects the regions with garments that are bottoms or tops. Our solution implements models that are capable of finding human parts in an image e.g. full-body vs half-body, or no human is found. Then, other models knowing that there's a human and its composition (e.g. not always we have a full-body) finds the bounding boxes/regions of the image that very likely correspond to a bottom or a top. For the creation of bounding boxes/regions task, a benchmark dataset was specifically prepared. The results show that the Mask RCNN solution is robust, and generalized enough to be used and scalable in unseen apparel/fashion data. 
\end{abstract}

\section{Introduction}\label{sec:intro}
Fashion industry valued at 3 trillion dollars \cite{global} is an industry with thousands of images taken or uploaded daily. Computer Vision techniques are being widely used in the industry \cite{cheng2021fashion} for Fashion Detection \cite{kim2021multiple,rame2020core,uzum2021deep}, Analysis \cite{liu2021mmfashion,gu2020fashion,han2021color,zhang2022aablstm}, Synthesis \cite{pandey2020poly,chen2020tailorgan,kwon2022tailor} and Recommendation \cite{ravi2021buy,zheng2021personalized,tangseng2020toward,10.1007/978-3-030-49186-4_36}. Instance Segmentation can be used in fashion images when is needed to isolate individual garments from outfits.

Images may contain only a single garment or multiple garments depending on the source or the purpose of the image. The images are generally noisy, with different backgrounds, positions and contain one or more garments. Basically, it is a very heterogenous space and, consequently, this work aims to introduce an automated way to apply Computer Vision architecture. The first step in the proposed solution starts with an image classification model to classify fashion images into five classes with the objective of separating images that show a single garment from images that show multiple garments. Additionally, we develop an instance segmentation model to segment the images with more than one garment in order to detect top and bottom garments. Each garment will have four outputs: first, a segmentation mask that surrounds the segmented object; second, a bounding box that contains the garment found in image; a class that determines if the garment is top or bottom, and, finally, a confidence score associated with the models classification.

Fashion datasets that are made available \cite{zheng2018modanet, jia2020fashionpedia,ge2019deepfashion2,liu2016deepfashion} usually contain many classes, which can have higher or lower granularity about the information related to the garments. Although this is very useful information for many specific problems where those classes labels are highly relevant, what we found missing was a model that is able to identify from a fashion image the section that contains only a top or bottom garment. 

This paper introduces a solution that classifies the human body completeness (full body, half body) and if contains only the garment the label for top or bottom. Additionally, in our training data we have images labeled as "accessories" and those do not need to go through the segmentation step to find the body completeness, nor the garment top/bottom information.

As a final component of our proposed approach is an instance segmentation dataset that uses polygons to properly identify the location of a top or a bottom within an image. One of the contributions of this work will be a dataset \footnote{The dataset is available upon request} that allows to train a Mask RCNN model for instance segmentation on fashion images. 

The remainder of the paper is organized as follows. In Section~\ref{sec:related} we give an overall review on the related work that has been done. In Section~\ref{sec:dataset} an explanation of the data used is given. The proposed dataset is explained with more details. In Section~\ref{sec:implementation}, we describe how we implemented our pipeline and in Section~\ref{sec:results} we talk about the results we obtained. Finally, in Section~\ref{sec:conclusion} our work is concluded and directions for future work are proposed.

\begin{figure*}[ht]
  \centering
  \includegraphics[width=0.9\linewidth, height=2.5cm]{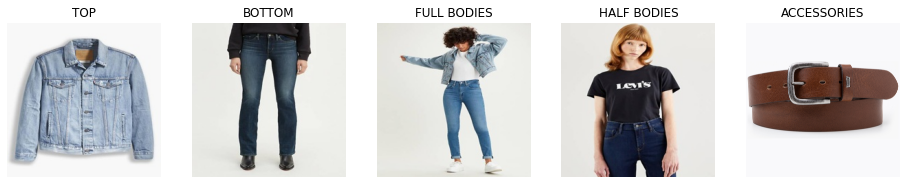}
  \caption{An example image of each class.}
  % \Description{Image contains a top, a bottom, a full body, a half body and a belt (accessory)}
  \label{fig:5classes}
\end{figure*}

\section{Related Work}\label{sec:related}
In \textit{CVPR 2020 Workshop}, DeepFashion 2 Challenge \footnote{https://sites.google.com/view/cvcreative2020/deepfashion2} was proposed based on DeepFashion1 \cite{liu2016deepfashion} and DeepFashion2 \cite{ge2019deepfashion2} datasets. EBay presented a new database called "ModaNet" \cite{zheng2018modanet}, aiming to provide a technical benchmark in evaluating the application of Computer Vision techniques for fashion understanding. Fashionpedia \cite{jia2020fashionpedia} proposed an ontology built by fashion experts and a dataset, consisting of annotated images taken from everyday and celebrity events.

Regarding instance segmentation, work has been done in two different paths, R-CNN and YOLO networks. Girschick et al proposed R-CNN, Regions with CNN features \cite{girshick2014rich}, an object detection algorithm that achieves a mean average precision (mAP) of 53.3\% on VOC 2012 \cite{everingham2011pascal}. R-CNN consists of a module that generates category independent region proposals, a large CNN that extracts a fixed-length feature vector from each region and a set of class-specific linear SVMs. Girschick extended it in \cite{girshick2015fast} and proposed Fast R-CNN, a faster method that achieves better results. Multiple regions of interest along with the image are input into the network. The network produces a convolutional feature map, extracts a feature vector for each object proposal and has as output probability estimates over all candidate classes and a bounding box for all the classes. Ren et al introduced Faster R-CNN \cite{ren2015faster}, a system with a Region Proposal Network, that is trained to generate region proposals, and a Fast R-CNN where the proposals are then input for detection. Mask R-CNN \cite{he2017mask} consists of a branch that predicts a segmentation mask, as well as a branch for bounding box recognition. Mask R-CNN can be generalized for human pose estimation and person keypoint detection. It outperforms all existing systems in all tasks.

YOLO networks were first presented by Redmon et al in \cite{redmon2016you} for object detection in real time. Compared to other state of the art object detection systems at the time it was presented, YOLO makes more errors but runs extremely fast.

\section{Dataset}\label{sec:dataset}
For the implementation of the pipeline and the creation of the dataset, images from DeepFashion1 dataset proposed by Liu et al \cite{liu2016deepfashion} and an apparel company are being used. The reason is that we aimed to have images from different datasets to generalise the pipeline to be able to segment more images.

For training and testing an instance segmentation model to detect top and bottom garments in images with outfits a fashion dataset annotated with polygons on tops and bottoms is needed. However, most fashion datasets proposed for instance segmentation contain more classes than the two needed and are over complicated for our task. DeepFashion2 \cite{ge2019deepfashion2} and ModaNet \cite{zheng2018modanet} datasets contain 13 classes (e.g. bag, belt, boots, etc), while Fashionpedia \cite{jia2020fashionpedia} contains 27 (e.g. collar, sleve, pocket, etc). 

\begin{figure}[ht]
  \centering
  \begin{subfigure}{0.45\linewidth}
    \includegraphics[width=0.6\linewidth]{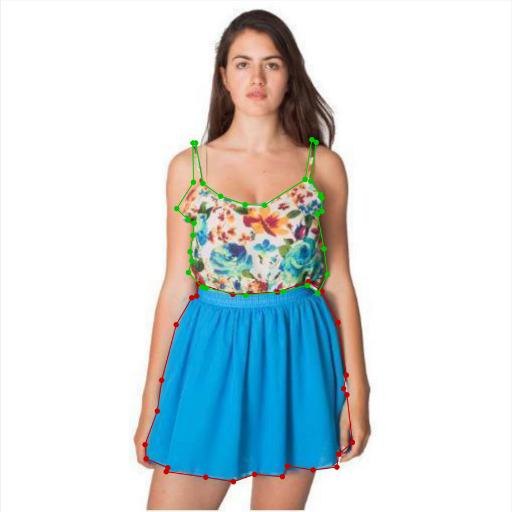}    
    \label{fig:dataset_poly-a}
  \end{subfigure}
  \hfill
  \begin{subfigure}{0.45\linewidth}
    \includegraphics[width=0.6\linewidth]{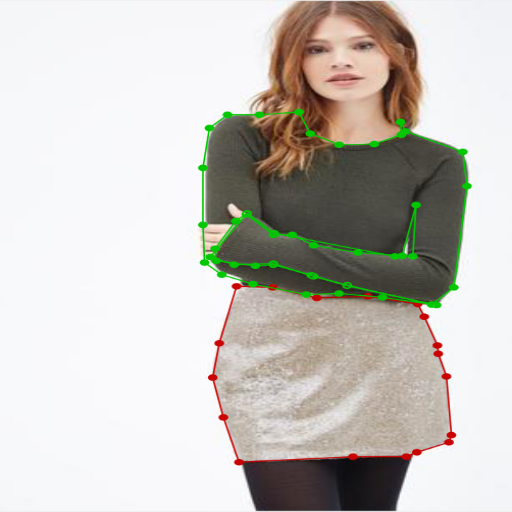}
    \label{fig:dataset_poly-b}
  \end{subfigure}
  \caption{Dataset with polygons.}
  % \Description{Two images of outfits that are manually annotated using polygons. Top garments are annotated with a green polygon and bottom garments are annotated with a red polygon.}
  \label{fig:dataset_poly}
\end{figure}

With the use of LabelMe tool \cite{russell2008labelme}, we draw polygons for tops and bottoms on images. It was chosen 320 images displaying clothes on humans from both datasets. Our goal is to create a generic model, hence it is included a diverse set of human poses e.g. front or arms folded and images with monochrome outfits for better segmentation. Every image contains at most 2 objects, a top and a bottom garment respectively. Examples of images with polygons from our dataset are shown in Figure~\ref{fig:dataset_poly}. Green polygons (the object above) indicate a top garment and red polygons (the object below) indicate a bottom garment.

The annotated images will be used for training, testing and validating the model, without overlapping between the three.

\section{Implementation}\label{sec:implementation}
\begin{figure*}[ht]
  \centering
  \includegraphics[width=0.6\linewidth, height=8cm]{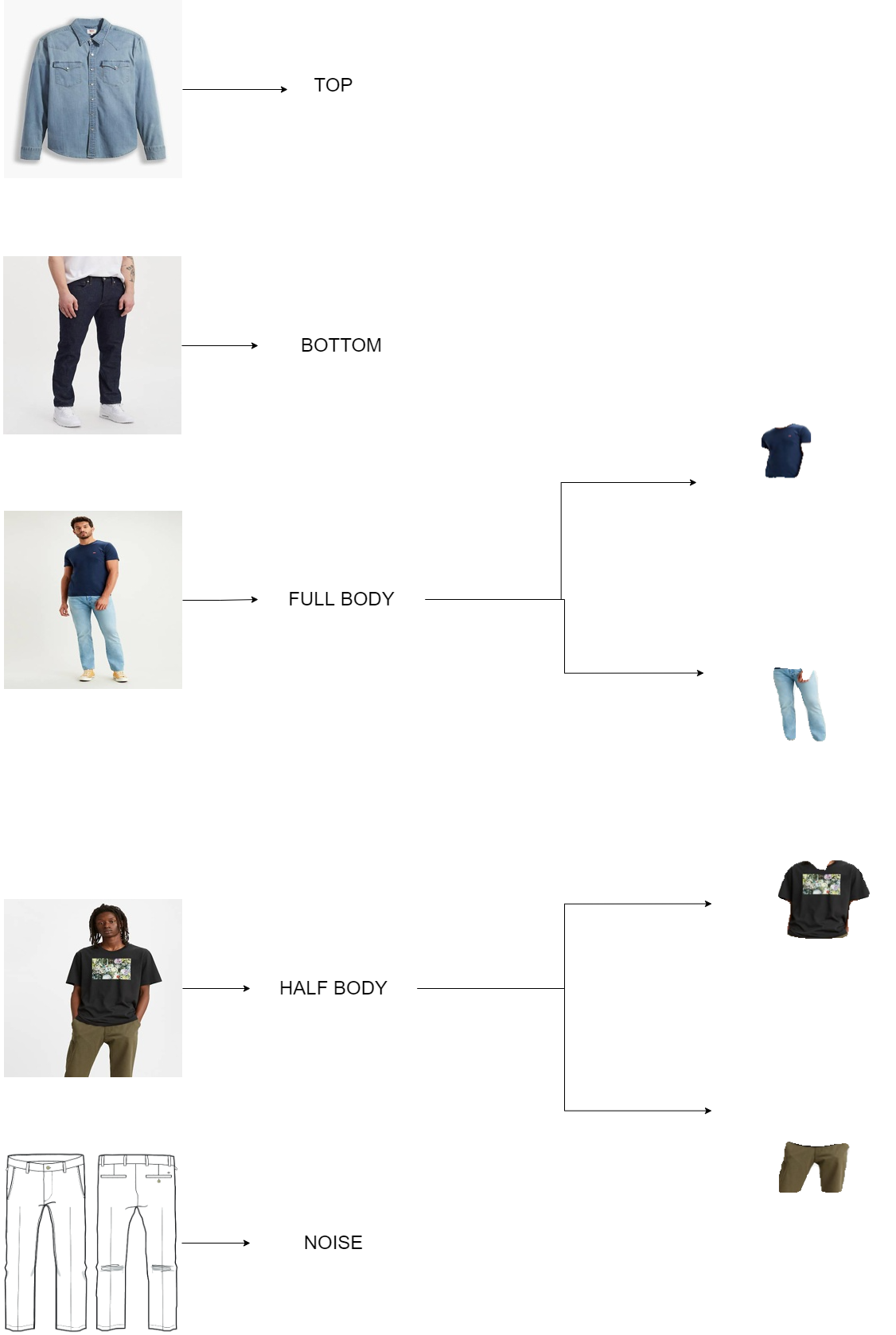}
  \caption{Images are first classified and then segmented.}
  % \Description{}
  \label{fig:data_flow}
\end{figure*}

The work we propose in this paper is separated in two different models, image classification and instance segmentation. It is divided into two models as the objective is to only segment images that show more than one garment. Images classified as tops or bottoms do not need to be segmented as they already show exactly one garment. On the contrary, if the image contains both a top and a bottom, then it needs to be segmented to recognize the individual garment. 
In Figure~\ref{fig:data_flow} a data flow of our pipeline is shown. Images are first classified into top, bottom, full body, half body or noise and images classified as full body or half body are then segmented to detect a top and a bottom garment. In the next sections a more detailed explanation on the work that has been done for the two models will be given.

\subsection{Image Classification}

Detecting tops and bottoms from images is a task that requires a few steps to be done prior. In the proposed work it is required to first obtain only images that contain both top and bottom garments. Therefore, we use image classification to get those images. With the use of BodyPix API that uses ResNet50 \cite{he2016deep} as backbone network, we segment the human or the garment from the background of the image. Once the background is removed there are smaller objects that remain in the image. As a strategy to the problem the algorithm considers the largest component in the image to be the only necessary and it removes all the smaller components in the image. Subsequently, it is checked for images that have been destroyed (the main object is removed from the image) during the process in order to remove them. Then, the data are separated in three different classes, with a rather broad classification. If a top or bottom garment is fully shown with a small part of the other garment, the image is classified as either top or bottom depending on the garment that is fully shown. If both a top and a bottom are shown in the image not necessarily completely, then the image is classified as Full Body.

The model is fine tuned on three different models to see the behavior on our task. Pre-trained weights of ResNet50 \cite{he2016deep}, VGG16 \cite{simonyan2014very} and InceptionV3 \cite{szegedy2016rethinking} models are used, all trained on ImageNet \cite{deng2009imagenet} dataset. From the visual results, we notice that the model didn't behave as expected for images where humans were cropped around the knees. Some of those images were classified as Tops and some images were classified as Full Bodies.

It is decided to include another class, named Half Bodies in the model to make it operate better. The new class will have all images consisting humans that are cropped around the knees. Simultaneously, we decide to include a fifth class named Accessories to classify images that do not belong to any of the aforementioned classes. These images may illustrate accessories, shoes, garment drawings. Sometimes if a product is not made yet, instead of the image of the product we get a drawing of the product. However, the drawing as it is not the exact product, is classified in the Accessories class. Since we only want to detect tops and bottoms, images classified as Accessories do not need to be segmented. 

The five classes are described in Table~\ref{tab:5class_descriptions}. The three aforementioned pre-trained models are fine tuned using the data described in Section~\ref{sec:dataset} and the comparative accuracy results are reported in Table~\ref{tab:classifier_comparison} after 20 epochs. InceptionV3 model is the most appropriate among the pre-trained models to be used in our task. Even for that small amount of epochs, the difference between the models was significant. Therefore, we continue working only with InceptionV3, to achieve the best possible results. The used training parameters are shown in Table~\ref{tab:training_parameters}. 
The same process is repeated, but this time we remove the Accessories class and we train a four classes model. Our motive is that some datasets may not contain accessories garment drawings and therefore a four class model will be more appropriate to be used.

\begin{table*}
  \caption{Description of each of the five classes.}
  \label{tab:5class_descriptions}
  \begin{tabular}{ll}
    \toprule
    Class & Description \\
    \midrule
    Top & Only a piece of top is shown \\
    Bottom & Only a piece of bottom is shown \\
    Full Body & Both a top and a bottom piece are clearly shown \\
    Half Body & A top is clearly shown and a bottom is cut \\
    Accessories & Accessories, shoes, garment drawings, blank images \\
    \bottomrule
  \end{tabular}
\end{table*}

\begin{table}
  \caption{Accuracy of pre-trained models.}
  \label{tab:classifier_comparison}
  \begin{tabular}{lccc}
    \toprule
     & ResNet50 & VGG16 & InceptionV3 \\
    \midrule
    Accuracy & 69\% & 76\% & 86\% \\
    \bottomrule
  \end{tabular}
\end{table}

\begin{table}
  \caption{Training parameters for Image Classification.}
  \label{tab:training_parameters}
  \begin{tabular}{lc}
    \toprule
    Parameter & Value \\
    \midrule
    Learning Rate & 0.0001 \\
    Batch Size & 32 \\
    Loss Function & Categorical Cross Entropy \\
    Optimizer & Adam Optimizer \\
    % Training Images & 2000 \\
    % Validation Images & 1000 \\
    Epochs & 500 \\
    \bottomrule
  \end{tabular}
\end{table}

\subsection{Instance Segmentation}
After we classified the images, we obtained full and half body images. We now need to segment them to detect top and bottom garments. At that step, we want to obtain a segmentation mask and a bounding box for each object of the input images. As explained in 
Section~\ref{sec:related}, the two candidates to be used for instance segmentation are RCNN and YOLO models. We choose to use R-CNN as higher accuracy is preferred over real time segmentation in our task. Out of RCNN models, Mask RCNN is chosen as it outperforms previous versions of RCNN and it gives a mask as an output that could be useful for our task.

% We begin by using Pixellib library \cite{Olafenwa2021SimplifyingOS}. We fine tune using our custom dataset with polygons. The visual output is satisfying. However, the output we get using this library is an image with the masks drawn on the segmented objects for each input image. We search for an alternative with which we can obtain the masks as an output. 
% 
Matterport Mask RCNN implementation \cite{matterport_maskrcnn_2017} is used with pre-trained weights on ModaNet \cite{zheng2018modanet} and we fine tune using the dataset we introduced. The model is fine tuned only on head layers (all layers but the backbone) for the first epochs and on all layers for the rest of the epochs as described in Table~\ref{tab:training_parameters_is}. 

We want the model to be able to detect top and bottom garments, even if the image is noisy, blurred or contains monochrome outfits as they are common cases. Rather uncommon special cases that we also want to avoid are when the image is either flipped or rotated. Therefore we apply augmentation (rotation, flip and blur) in our images to improve the behavior of the model in those cases.

\begin{table}
  \caption{Training parameters for Mask RCNN model.}
  \label{tab:training_parameters_is}
  \begin{tabular}{lc}
    \toprule
    Parameter & Value \\
    \midrule
    Learning Rate (Head Layers) & 0.001 \\
    Learning Rate (All Layers) & 0.0001 \\
    Epochs (Head Layers) & 5 \\
    Epochs (All Layers) & 35 \\
    Learning Momentum & 0.9 \\
    Weight Decay & 0.0001 \\
    Backbone & ResNet101 \\
    % Steps per Epoch & 240 \\
    % Validation Steps & 60 \\
    % Training Images & 240 \\
    % Validation Images & 60 \\
    \bottomrule
  \end{tabular}
\end{table}

\section{Results}\label{sec:results}
The first model is tested with 100 images taken from the datasets explained in Section~\ref{sec:dataset}, 20 of each class for the 5 class model and 25 of each class for the 4 class model. In Table~\ref{tab:classification_report5} and Table~\ref{tab:classification_report4} we present the classification report for the 5 and 4 class models. The 4 class model has a 3 \% higher overall accuracy than the 5 classes model. The 5 class model has more complexity so it is expected to perform slightly worse than the 4 class model. If the dataset allows it, then the 4 class model should be used.

The second model is tested using 20 images, from the dataset we introduced. The images used for testing are different from the images used during the training phase. Various Intersection over Union (IoU) thresholds are set and we get the mean Average Precision (mAP) for each IoU threshold. Also tests are conducted using different pre-trained weights to see how they operate in our task. We use pre-trained weights from Imagenet \cite{deng2009imagenet}, COCO \cite{lin2014microsoft} and ModaNet \cite{zheng2018modanet} and we report the mAP for various IoU thresholds after fine tuning for 10 epochs in Table~\ref{tab:comparison_segm}. We include the model we used (ModaNet(40)) in the table for comparison. ModaNet weights achieve significantly better mAP than Imagenet and COCO weights under all IoU thresholds. For IoU threshold lower than 0.8, ModaNet achieves high mAP. '

\begin{figure*}
  \centering
  \begin{subfigure}{0.9\linewidth}
    \includegraphics[width=0.95\linewidth]{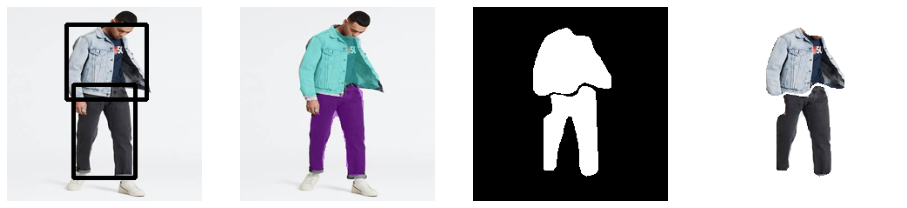}
    \label{fig:segmentedA}
  \end{subfigure}
  \begin{subfigure}{0.9\linewidth}
    \includegraphics[width=0.95\linewidth]{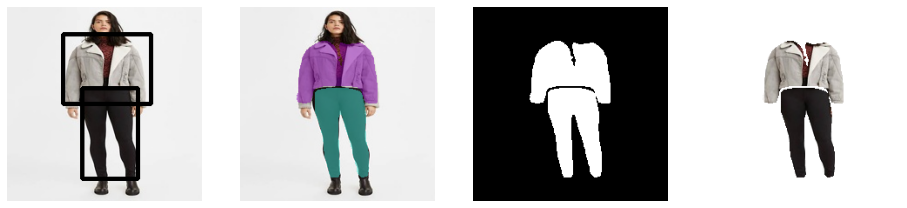}
    \label{fig:segmentedB}
  \end{subfigure}
  % \Description{}
  \caption{Images segmented.}
  \label{fig:segmented}
\end{figure*}

\begin{figure}
%   \centering
  \begin{subfigure}{0.45\linewidth}
  \centering\captionsetup{width=.6\textwidth}%
    \includegraphics[width=0.6\linewidth, height=2.5cm]{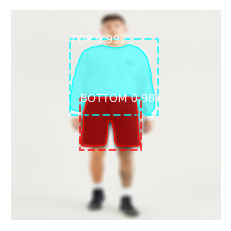}
    \caption{Blurred image.}
    \label{fig:special_blur}
  \end{subfigure}
  \hfill
  \begin{subfigure}{0.45\linewidth}
  \centering\captionsetup{width=.6\linewidth}%
    \includegraphics[width=0.6\linewidth, height=2.5cm]{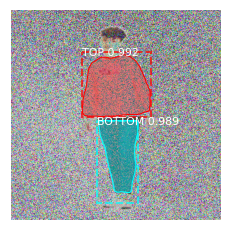}
    \caption{Image with added noise.}
    \label{fig:special_noise}
  \end{subfigure}
  \hfill
  \begin{subfigure}{0.45\linewidth}
  \centering\captionsetup{width=.6\linewidth}%
    \includegraphics[width=0.6\linewidth, height=2.5cm]{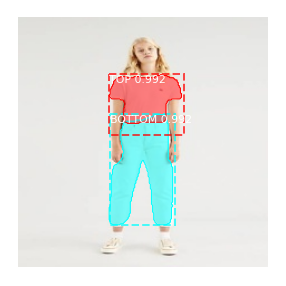}
    \caption{White monochrome image.}
    \label{fig:special_monochrome}
  \end{subfigure}
  \hfill
  \begin{subfigure}{0.45\linewidth}
  \centering\captionsetup{width=.6\linewidth}%
    \includegraphics[width=0.6\linewidth, height=2.5cm]{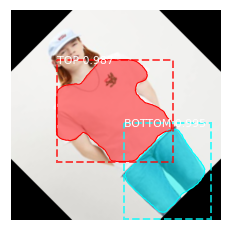}
    \caption{Image rotated 45 degrees.}
    \label{fig:special_rotate45}
  \end{subfigure}
  \hfill
  \begin{subfigure}{0.45\linewidth}
  \centering\captionsetup{width=.6\linewidth}%
    \includegraphics[width=0.6\linewidth, height=2.5cm]{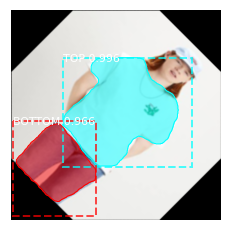}
    \caption{Image rotated 315 degrees.}
    \label{fig:special_rotate315}
  \end{subfigure}
  \hfill
  \begin{subfigure}{0.45\linewidth}
  \centering\captionsetup{width=.6\linewidth}%
    \includegraphics[width=0.6\linewidth, height=2.5cm]{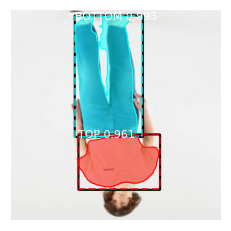}
    \caption{Flipped image.}
    \label{fig:special_flip}
  \end{subfigure}
  % \Description{}
  \caption{Test on special cases.}
  \label{fig:special_cases}
\end{figure}

The output of the segmentation for two images is presented in Figure~\ref{fig:segmented}. For each input image we obtain the bounding box and segmentation mask for each detected garment. With those outputs, we can isolate the apparel off the human. We test the model for special cases to check the robustness of the model and we present the results in Figure~\ref{fig:special_cases}.

In Figure~\ref{fig:special_blur} a blurred images and in Figure~\ref{fig:special_noise} a noisy image are entered as input. Figure~\ref{fig:special_monochrome}, shows an image of a monochrome white outfit that has been segmented by the model. In Figure~\ref{fig:special_rotate45} and Figure~\ref{fig:special_rotate315} we present the output of the model when the input image is rotated and Figure~\ref{fig:special_flip} displays the output of the model when the input image is flipped. 

\begin{table}
  \caption{Image classifier [5 classes]}
  \label{tab:classification_report5}
  \begin{tabular}{lcccc}
    \toprule
      & Precision & Recall & F1-Score & Support \\
    \midrule
    Tops & 1.00 & 0.95 & 0.97 & 20 \\
    Bottoms & 0.65 & 1.00 & 0.78 & 20 \\
    Full Bodies & 0.93 & 0.65 & 0.76 & 20 \\
    Half Bodies & 0.94 & 0.75 & 0.83 & 20 \\
    Accesories & 0.95 & 0.95 & 0.95 & 20 \\
    Accuracy &  &  & 0.86 & 100\\
    \bottomrule
  \end{tabular}
\end{table}

\begin{table}
  \caption{Image Classifier [4 classes]}
  \label{tab:classification_report4}
  \begin{tabular}{lcccc}
    \toprule
      & Precision & Recall & F1-Score & Support \\
    \midrule
    Tops & 0.92 & 0.92 & 0.94 & 25 \\
    Bottoms & 0.86 & 1.00 & 0.93 & 25 \\
    Full Bodies & 0.88 & 0.88 & 0.88 & 25 \\
    Half Bodies & 0.90 & 0.72 & 0.80 & 25 \\
    Accuracy &  &  & 0.89 & 100\\
    \bottomrule
  \end{tabular}
\end{table}

\begin{table*}
  \caption{mAP comparison}
  \label{tab:comparison_segm}
  \begin{tabular}{lccccc}
    \toprule
     & mAP50 & mAP60 & mAP70 & mAP80 & mAP90\\
    \midrule
    Imagenet& 0.5 & 0.38 & 0.13 & 0.05 & 0.0\\
    COCO& 0.88 & 0.73 & 0.63 & 0.27 & 0.1\\
    ModaNet(10)& 0.95 & 0.9 & 0.85 & 0.75 & 0.25\\
    ModaNet(40)& 0.975 & 0.975 & 0.925 & 0.75 & 0.438\\
    \bottomrule
  \end{tabular}
\end{table*}

% \begin{figure}
%   \centering
%   \begin{subfigure}{0.45\linewidth}
%   \includegraphics[width=0.9\linewidth]{images/Confusion Matrix 5 classes.png}
%     \caption{Confusion matrix for 5 classes model.}
%     \label{fig:confusion5}
%   \end{subfigure}
%   \hfill
%   \begin{subfigure}{0.45\linewidth}
%   \includegraphics[width=0.9\linewidth]{images/Confusion Matrix 4 classes.png}
%   \caption{Confusion matrix for 4 classes model.}
%   \label{fig:confusion4}
%   \end{subfigure}
%   \Description{}
%   \caption{Confusion matrices.}
%   \label{fig:confusion}
% \end{figure}

\section{Conclusion and future work}\label{sec:conclusion}
In this paper, we propose a pipeline consisting of two models for detecting tops and bottoms out of fashion images. The first model classifies the images in five classes and the second model segments the images classified in two of those five classes to detect top and bottom garments. The pipeline is robust and capable to operate for extreme inputs such as monochrome outfits, blurred, noisy, rotated or flipped images. It is introduced a dataset consisting of 320 fashion images, each of them labelled with polygon annotations. All images show outfits on humans with two annotated objects, a top and a bottom garment. The dataset can be used for instance segmentation on fashion images. Possible improvements of our work include the strengthen of the model to operate with more types of garments, to be capable of detecting the specific product for each segmented object, while also further expanding the dataset.

\bibliographystyle{ieee_fullname}
\bibliography{references}
\end{document}